\documentclass[letterpaper]{article} 
\usepackage{aaai24}  
\usepackage{times}  
\usepackage{helvet}  
\usepackage{courier}  
\usepackage[hyphens]{url}  
\usepackage{graphicx} 
\usepackage{natbib}  
\usepackage{caption} 
\usepackage{algorithm}
\usepackage{algorithmic}

\usepackage{amsfonts}
\usepackage{bm}
\usepackage{amsmath} %
\usepackage{booktabs}
\usepackage{multirow}
\usepackage{colortbl}
\usepackage{tabularx}
\usepackage{amssymb}

\usepackage{newfloat}
\usepackage{listings}
\DeclareCaptionStyle{ruled}{labelfont=normalfont,labelsep=colon,strut=off} 
\floatstyle{ruled}
\newfloat{listing}{tb}{lst}{}
\floatname{listing}{Listing}
\title{TR-DETR: Task-Reciprocal Transformer for Joint Moment Retrieval and Highlight Detection}

\author{
    Hao Sun\textsuperscript{\rm 1,2,3}\equalcontrib,
    Mingyao Zhou\textsuperscript{\rm 1,2,3}\equalcontrib,
    Wenjing Chen\textsuperscript{\rm 4}\thanks{Corresponding authors.},
    Wei Xie\textsuperscript{\rm 1,2,3}\footnotemark[2]
}
\affiliations{
    \textsuperscript{\rm 1}Hubei Provincial Key Laboratory of Artificial Intelligence and Smart Learning,\\ Central China Normal University, Wuhan, China \\
    \textsuperscript{\rm 2}School of Computer Science, Central China Normal University, Wuhan, China \\
    \textsuperscript{\rm 3}National Language Resources Monitoring and Research Center for Network Media, \\Central China Normal University, Wuhan, China \\
    \textsuperscript{\rm 4}School of Computer Science, Hubei University of Technology, Wuhan, China \\
    haosun@ccnu.edu.cn, zhoumingyao@mails.ccnu.edu.cn, chenwenjing@hbut.edu.cn, XW@mail.ccnu.edu.cn
}

\begin{document}

\maketitle

\begin{abstract}
    Video moment retrieval (MR) and highlight detection (HD) based on natural language queries are two highly related tasks, which aim to obtain relevant moments within videos and highlight scores of each video clip. Recently, several methods have been devoted to building DETR-based networks to solve both MR and HD jointly. These methods simply add two separate task heads after multi-modal feature extraction and feature interaction, achieving good performance. Nevertheless, these approaches underutilize the reciprocal relationship between two tasks. In this paper, we propose a task-reciprocal transformer based on DETR (TR-DETR) that focuses on exploring the inherent reciprocity between MR and HD. Specifically, a local-global multi-modal alignment module is first built to align features from diverse modalities into a shared latent space. Subsequently, a visual feature refinement is designed to eliminate query-irrelevant information from visual features for modal interaction. Finally, a task cooperation module is constructed to refine the retrieval pipeline and the highlight score prediction process by utilizing the reciprocity between MR and HD. Comprehensive experiments on QVHighlights, Charades-STA and TVSum datasets demonstrate that TR-DETR outperforms existing state-of-the-art methods. Codes are available at \url{https://github.com/mingyao1120/TR-DETR}.\footnote{Note: This is a pre-print version of the paper. The final, copyrighted version of the paper can be accessed through the AAAI Digital Library.}
\end{abstract}

\section{Introduction}
With the ubiquity of digital devices and the expansion of the Internet, the number and variety of videos are rapidly increasing~\cite{Foo_2023_CVPR}. How to quickly search out the desired moments from massive videos (called moment retrieval, MR)~\cite{gao2017tall} and efficiently browse videos (called highlight detection, HD)~\cite{DBLP:conf/mm/MolinoG18} according to the needs of users has attracted widespread attention. In practical applications, user needs can be expressed in natural language queries~\cite{DBLP:conf/aaai/00010WLW22}. Due to the complexity of video content as well as the diversity of user needs, MR\&HD based on user-provided natural language queries is extremely challenging.

The goal of MR is to precisely search for semantically related moments from whole videos guided by natural language queries~\cite{DBLP:conf/cvpr/0006XQZT0YZW22}. The common pipeline of MR involves several steps. Firstly, pre-trained networks are utilized to extract features from the input video and text. Subsequently, cross-modal interaction is performed based on the extracted features to obtain the query relevance score of the candidate moment or the frame-level start-end probability of the relevant moment~\cite{DBLP:journals/pami/ZhangSJZ23}. HD based on queries strives to assign highlight scores to each video clip based on considering the user needs~\cite{DBLP:journals/tmm/GuoZJWLY22}. Existing methods~\cite{liu2022umt,xiong2023dual} utilize transformers~\cite{vaswani2017attention} or graph neural networks~\cite{scarselli2008graph} to perform single-modal feature encoding or cross-modal interaction.

Due to the task similarity between MR and HD based on queries, and the commonality between their methods involving multi-modal feature extraction, feature interaction, etc., some works~\cite{lei2021detecting, DBLP:journals/corr/abs-2307-16715} have devoted to designing various multi-task networks for joint MR\&HD. For example, Moment-DETR~\cite{lei2021detecting} pioneers the application of DETR~\cite{DBLP:conf/eccv/CarionMSUKZ20} for joint MR\&HD. QD-DETR~\cite{moon2023query} introduces a query-dependent video representation module, making moment predictions reliant on user queries. MH-DETR~\cite{xu2023mh} introduces a pooling operation into the encoder and incorporates a cross-modality interaction module to fuse visual and query features. In these methods, two isolated task heads are added after the shared multi-modal feature extraction and feature interaction modules for joint MR\&HD.
These methods generally focus on improving the discrimination of multi-modal feature extraction and feature interaction through a multi-task learning scheme, achieving good performance. However, the reciprocity between MR and HD tasks is ignored.

For MR, the highlight scores from HD based on user-provided queries can be utilized to assist in eliminating query-irrelevant clips, thereby boosting moment retrieval accuracy. In turn, for HD based on queries, the results of moment retrieval can be used to improve the understanding of videos and user needs. Therefore, MR and HD based on queries are reciprocal.

To fully exploit the reciprocal relationship between the two tasks, we propose a task-reciprocal transformer based on DETR, named TR-DETR, for joint MR\&HD. Firstly, visual features and textual features are extracted from user-provided videos and queries through pre-trained networks. Then, we introduce a local-global multi-modal alignment module to perform local and global semantic alignment before modal interaction, respectively. This module encourages the model to distinguish video clips that are semantically similar but irrelevant to the query. Subsequently, we propose a visual feature refinement module for modal interaction, which employs aligned textual features to filter out query-irrelevant information in visual features to avoid it interfering with joint features. Finally, to exploit the complementarities between MR and HD, we propose a task cooperation module consisting of HD2MR and MR2HD. The former explicitly infuses highlight score information into the moment retrieval process, enhancing localization accuracy. The latter exploits localization outcomes to derive clip-level relevant scores, offering visual support for highlight detection.
Extensive experiments on QVHighlights~\cite{lei2021detecting}, Charades-STA~\cite{gao2017tall} and TVSum~\cite{song2015tvsum} demonstrate that the proposed TR-DETR outperforms the state-of-the-art methods. The contributions of this paper are summarized as follows:

\begin{itemize}
    \item We highlight the reciprocity between MR and HD. In addition, we introduce an innovative TR-DETR network that leverages this reciprocity between tasks to optimize performance.
    \item We introduce the local and global alignment regulators. These regulators are designed to facilitate semantic alignment between video clips and the query, which serves to generate discriminative joint representations.
    \item To explore the intrinsic complementarity between the two tasks, we construct a task cooperation module. This module explicitly exploits the complementarity between MR and HD by injecting highlight scores into the moment retrieval pipeline and using the retrieved moments to refine the initial highlight distribution.
\end{itemize}

\begin{figure*}[t]
\centering
\includegraphics[width=1\textwidth]{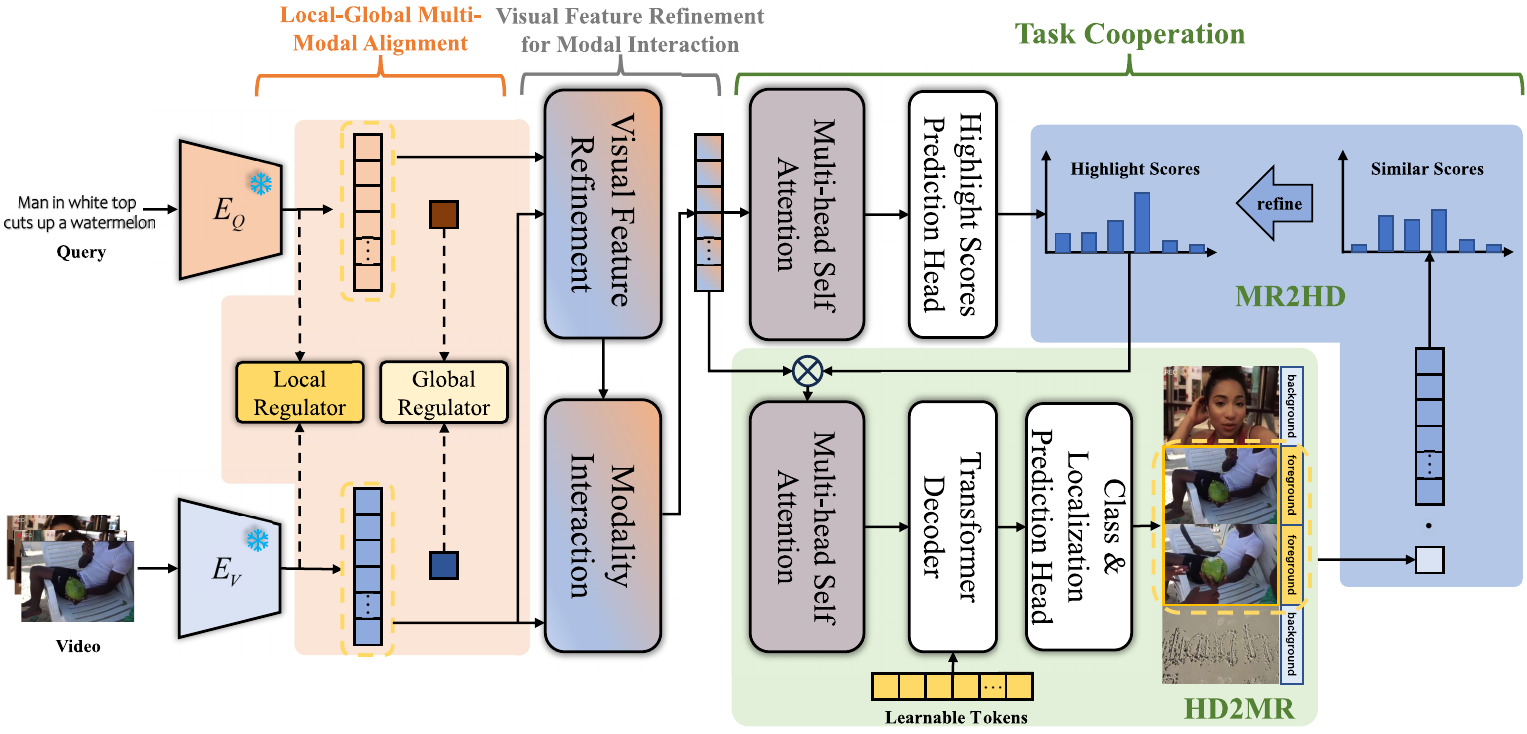} 
\caption{The proposed TR-DETR involves several key steps. Initially, two frozen pre-trained networks are employed to extract visual and textual features from videos and queries. Subsequently, a local-global multi-modal alignment module is constructed to effectively align the extracted visual and textual features. Then, the visual features are refined under the guidance of textual features for obtaining discriminative joint features. Finally, a task cooperation module is implemented to enhance prediction outcomes based on task reciprocity. Additionally, two multi-head self-attention components share weights.}
\label{method}
\end{figure*}

\section{Related Works}
\subsection{MR and HD}

Video moment retrieval is originally introduced by the literature~\cite{gao2017tall}, with the objective of retrieving moments from a video based on a given natural language query. Moment retrieval typically includes two types of methods: proposal-based and proposal-free methods. In the proposal-based methods, candidate moments are initially generated through techniques such as sliding windows~\cite{gao2017tall}, proposal generation networks~\cite{xu2019multilevel}, or 2D-Maps~\cite{zhang2020learning}. These candidates are subsequently ranked based on the similarity scores to the query, where the candidate with the highest score is used as the result. Although these methods have high accuracy, they necessitate additional pre- and post-processing steps, introducing computational redundancy. Moreover, their performance heavily relies on the quality of candidate moments. On the other hand, proposal-free methods~\cite{ghosh2019excl, zhang2021natural,mun2020local} directly predict start-end probabilities for target moments within a video, which eliminates the need to rank a large number of candidate moments, thereby improving training efficiency.

In contrast, highlight detection concentrates on measuring the significance of each clip within a given video. Slightly different from moment retrieval, highlight detection initially is proposed as a single-modal task and does not rely on text queries. However, highlight determination is often a subjective matter and users' preferences should be taken into account. Therefore, the literature~\cite{kudi2017words} proposes to integrate text queries as supplementary information for highlight detection. Nonetheless, this work relies solely on text ranking algorithms to rank video descriptions in the text domain to guide video clip ranking. It does not entail a direct alignment of text and highlights. Subsequently, in video thumbnail generation, which closely parallels highlight detection, Yuan \emph{et al.}~\cite{DBLP:conf/mm/YuanM019} delves into text queries and uses graph convolutional networks to model the interaction between each clip and text.

Conventionally, moment retrieval and highlight detection are addressed in isolation, lacking an integrated framework for joint learning. Recent research~\cite{lei2021detecting} constructs the QVHighlights dataset to facilitate joint learning of MR\&HD and proposes a baseline model based on DETR. Building upon this, Liu \emph{et al.}~\cite{liu2022umt} incorporates audio modality into the model, catering to scenarios for missing queries. Additionally, Moon \emph{et al.}~\cite{moon2023query} prioritizes full integration of provided query information into the joint representation, enabling the text to guide both moment retrieval and highlight detection. Different from previous methods, this paper focuses on exploiting the natural reciprocity between two tasks.

\subsection{Multi-Modal Alignment}
Recently, researchers in the multimodal field have focused on constructing contrastive losses to fit the interactions and correspondences between different modalities~\cite{luo2020univl,sun2019learning,miech2020end, yan2023CAAI}. For example, the literature~\cite{ging2020coot} introduces a cycle consistency loss to align video clip-level features and query word-level features. Similarly, the literature ~\cite{zhang2022video} introduces a multi-level contrast loss to capture multi-granular interactive alignment details within queries and videos, enhancing the performance of moment retrieval.
Although these methods share similarities with the multi-modal alignment in our approach, they do not explicitly align the semantic information of different modalities before modality interaction, resulting in insufficient discrimination of joint features.

\section{Method}
The overview of TR-DETR is shown in  Figure~\ref{method}.
TR-DETR comprises four core modules: feature extraction, local-global multi-modal alignment, visual feature refinement for modal interaction, and task cooperation.
Details are introduced as follows.

\subsection{Feature Extraction}
\subsubsection{Visual Features.}
Following the literature~\cite{lei2021detecting}, the video is first divided into non-overlapping clips according to a certain time interval, such as 2s. Then the pre-trained ViT-B/32 in CLIP~\cite{DBLP:conf/icml/RadfordKHRGASAM21} and SlowFast~\cite{DBLP:journals/corr/abs-1812-03982} are utilized to extract clip-level visual features $ F_{v}=\left[f_{v}^{1},f_{v}^{2},\ldots,f_{v}^{L}\right]\in \mathbb{R}^{L\times d_{v}}$, where $L$ and $d_v$ are the number of clips and the visual feature dimension, respectively. Following the way that UMT~\cite{liu2022umt} uses audio information, we use the pre-trained audio feature extractor to extract the audio features $F_a \in \mathbb{R}^{L\times d_{a}} $, and then splice them behind the visual features $F_{v}$. See the experimental settings for details.
\subsubsection{Textual Features.}
For a natural language query, we use the textual encoder in the pre-trained CLIP to extract textual features $ F_{t}=\left[f_{t}^{1},f_{t}^{2},\ldots,f_{t}^{N}\right]\in \mathbb{R}^{N\times d_{t}}$, where $N$ and $d_t$  are the number of words and the textual feature dimension, respectively.

\subsection{Local-Global Multi-Modal Alignment}
Existing methods~\cite{moon2023query,lei2021detecting,liu2022umt} for joint MR\&HD directly input the extracted visual and textual features into the modal interaction module to obtain joint features. However, there is a natural information mismatch between visual features and textual features, resulting in insufficient discrimination of joint features~\cite{10123038}. In this study, to reduce the modal gap, we propose a local-global multi-modal alignment module, comprising local and global regularization components. The local regulator helps the model distinguish semantically similar but undesired clips, while the global regulator ensures that both modalities share a unified semantic space. Integrating these alignment regulators can significantly promote multimodal associations and facilitate subsequent modal interactions.

Given the clip-level visual features  $F_v$ of the video and the word-level textual features $F_t$  of the query, we first map them into the same dimension $d$ by using three-layer multilayer perceptions (MLP).
\begin{align}
    \widehat{F}_v & = \text{MLP}_v(F_v), \\
    \widehat{F}_t & = \text{MLP}_t(F_t).
\end{align}
For the local regulator, we calculate the cosine similarity between each clip and each word by using the following formula, obtaining a similarity matrix $S_{loc} \in\mathbb{R}^{L\times N}$.
\begin{align}
    S_{loc} = \sigma \left( \frac{\widehat{F}_{v} \, \widehat{F}_{t}^{\mathsf{T}}}{\|\widehat{F}_{v}\|_{2} \|\widehat{F}_{t}\|_{2}} \right),
\end{align}
where $\sigma$ is the sigmoid function. We employ mean-pooling to get $\widehat{S}_{loc} = \text{MeanPooling}(S_{loc}) \in \mathbb{R}^L$,  which measures the similarity between each video clip and the global textual features. Then, a local regular loss $\mathcal{L}_{loc}$ is used to encourage distinguishing video clips that are irrelevant to the query.
\begin{align}
\begin{aligned}
    \mathcal{L}_{local} = -\sum_{i=1}^{L} \left( C^{i} \log(\widehat{S}^{i}_{loc}) + (1 - C^{i}) \log(1 - \widehat{S}^{i}_{loc}) \right),
\end{aligned}
\end{align}
where $\widehat{S}^i_{loc}$ is the similarity score between the $i$-th video clip and the global textual features, and $C^i$ indicates whether the $i$-th video clip and the query are actually relevant. Specifically, according to ground truth in MR, if the $i$-th clip is relevant to the query,  $C^i$ is 1, otherwise 0.
For the global regulator, a multi-modal contrastive loss~\cite{li2021align} is employed to promote the similarity of global representations of paired videos and queries.
\begin{align}
    \mathcal{L}_{global} = -\frac{1}{B}\sum_{i=1}^{B}\log\frac{\exp( (G_v^i) \, (G_t^i)^{\mathsf{T}})}{\sum_{i=1}^{B}\sum_{j=1}^{B}\exp((G_v^i) \, (G_t^j)^{\mathsf{T}})},
\end{align}
where $B$ is the batch size,  $G_v^i \in \mathbb{R}^d$ and $G_t^i \in \mathbb{R}^d$ are the global feature of the $i$-th video and the $i$-th query in a training batch, respectively. Specifically, $G_v^i$ is obtained by averaging all clip features $\widehat{F}_v$ within the $i$-th video, and $G_t^i$ is derived by averaging word-level features $\widehat{F}_t$ in the $i$-th query.

\subsection{Visual Feature Refinement for Modal Interaction}
The goal of modal interaction is to generate discriminative joint features from visual and textual features~\cite{lei2021detecting}, which play a key role in joint MR\&HD.
In the literature~\cite{lei2021detecting}, visual and textual features are simply concatenated for modal interaction. However, videos generally contain a large number of clips irrelevant to the textual query, which may cause the model to pay too much attention to these irrelevant contents, resulting in ignoring the really important clips.

To suppress the interference of query-irrelevant information in visual features, we introduce a query-guided visual feature refinement module inspired by the literature~\cite{xiong2016dynamic} for modal interaction. This module employs the textual query as a guide to refine clip-level visual features to effectively suppress irrelevant information present in the video and retain temporal cues. The similarity matrix between aligned clip-level visual features and word-level textual features is calculated as:
\begin{align}
    A = \frac{\text{Linear}(\widehat{F}_{v}) \, \text{Linear}(\widehat{F}_{t})^{\mathsf{T}}}{\sqrt{d}},
\end{align}
where $A \in \mathbb{R}^{L \times N}$ is the similarity matrix and Linear($\cdot$) represents the linear projection layer. Then the similarity matrix is used to weigh and sum the query and video features respectively to obtain preliminary refinement features.
\begin{align}
    F_{v2q} &= A_r \, \widehat{F}_t, \\
    F_{q2v} &= A_{r} \, A_{c}^{\mathsf{T}} \, \widehat{F}_{v},
\end{align}
where $A_r$ and $A_c$ represent the results after row softmax normalization and column softmax normalization of $A$, $F_{v2q}$ and $F_{q2v}$ are the clip-level textual features and word-level visual features, respectively. Finally, to further use text queries to optimize clip-level visual features $\widehat{F}_v$, we perform the following feature concatenation and obtain the final refined clip features $\overline{F}_v$ through linear projection.
\begin{align}
    F^{Cat}_v &= \left[\widehat{F}_v\| F_{v2q}\| \widehat{F}_v \odot F_{v2q}\| \widehat{F}_v \odot F_{q2v}\| F_t^G \right], \\
    \overline{F}_v &= \text{Linear}(F^{Cat}_v),
\end{align}
where $F_t^G \in \mathbb{R}^{L \times d}$ is a matrix formed by copying and splicing the text global features obtained through the pooling operation, $\left[ \cdot\|\cdot\right]$ means concatenation, and $\odot$ is the Hadamard product.
Then, modality fusion is performed using a cross-attention layer to further incorporate query features into the joint features, where textual features are from the refined clip feature $Q_{v} = \text{Linear}_q(\overline{F}_v)$, key and value features are from the textual features $K_t=\text{Linear}_k(\widehat{F}_t)$ and $V_t=\text{Linear}_v(\widehat{F}_t)$.
\begin{align}
    Z = \operatorname{Attention}(Q_v, K_t, V_t) = \text{Softmax}\left(\frac{Q_v K_t^{\mathsf{T}}}{\sqrt{d}}\right) V_t,
\end{align}
where $Z \in\mathbb{R}^{L\times d} $ represents joint features through modal interaction between refined visual features and textual features.

\subsection{Task Cooperation}
Although previous methods~\cite{lei2021detecting,liu2022umt,moon2023query}  have attempted to jointly solve MR and HD, these methods usually focus on optimizing the shared multi-modal feature extraction and feature interaction modules to improve the discrimination of joint features using a multi-task learning framework. However, the inherent complementarity between MR and HD tasks is underutilized.

In essence, video clips with high highlight scores are often strong candidates for MR. Because highlight-worthy clips tend to possess enhanced visual significance and attraction. Additionally, clips within the moment relevant to the current query probably cover the highlights, too.
This is because query-relevant moments also contain visual expressions of user needs, which helps to refine the highlight score distribution from the visual perspective. Given these insights, we propose a task cooperation module consisting of HD2MR and MR2HD components.

\subsubsection{HD2MR}
MR can leverage the highlight scores obtained by HD to empower the exclusion of irrelevant or less attractive video clips. We first use the multi-head attention mechanism and a linear layer to obtain clip-level highlight scores from the joint features $Z$.
\begin{align}
    H &= \text{Linear}(\text{MHA}(Z)),
\end{align}
where MHA($\cdot$) represents multi-head attention that is employed to model video temporal information and $H \in\mathbb{R}^L$ is the predicted highlight scores.

To filter out non-highlight information in $Z$ and explicitly inject highlight scores information into the MR pipeline, we multiply the clip-level highlight scores $H$ with the joint features $Z$  to obtain the enhanced joint features $\overline{Z} \in \mathbb{R}^{L \times d}$.  Then, $\overline{Z}$ is input into the MHA again for joint features encoding.
\begin{align}
\begin{aligned}
    \overline{Z} &= \text{Softmax}(H) \odot Z,\\
    \widehat{Z} &= \text{MHA}\left(Z + \overline{Z}\right),
\end{aligned}
\end{align}
where $\widehat{Z}$ is the joint features of the perceived highlight scores.
Finally, these enhanced features $\widehat{Z}$ are fed into the transformer decoder and prediction head from the literature~\cite{DBLP:conf/iclr/LiuLZYQSZZ22} to obtain the ultimate retrieved moments.

\subsubsection{MR2HD}
HD, in turn, gains a deeper understanding of video content and user needs by leveraging the text query and retrieved moments from MR. We employ the gated recurrent unit (GRU)~\cite{chung2014empirical} to effectively capture global information from the retrieved moments.
\begin{align}
\begin{aligned}
    F^M_v &= \text{GRU}(m),
\end{aligned}
\end{align}
 where $m$ represents the clip feature vectors in $\widehat{F}_v$ of the retrieved moments from HD2MR and $F^M_v \in \mathbb{R}^d$ is the global feature vector of these retrieved moments. To use the visual information of the retrieved moments to refine highlight scores prediction, we calculate similarity scores between $F^M_v$ and visual features $\widehat{F}_v$.
\begin{align}
\begin{aligned}
    S_{ref} &= \frac{F^M_v \, \widehat{F}_v^\mathsf{T}}{\|F^M_v\|_2 \, \|\widehat{F}_v\|_2},
\end{aligned}
\end{align}
 where $S_{ref}\in\mathbb{R}^L$ is the correlation between clips and $F^M_v$.
 The highlight score refinement process involves multiplying the clip-level correlation scores by $\widehat{Z}$, then adding them to $Z$, and finally obtaining refined scores by linear projection. The formulation is as follows:
 \begin{align}
    \begin{aligned}\overline{H}&=\text{Linear}(Z+\text{Softmax}(S_{ref})\odot \widehat{Z}),\end{aligned}
 \end{align}
where $\overline{H} \in\mathbb{R}^L$ is the refined highlight scores.

\subsection{Objective Losses}
The objective losses of TR-DETR include three parts: MR loss $\mathcal{L}_{mom}$, HD loss $\mathcal{L}_{high}$, regulators losses $\mathcal{L}_{local}$ and $\mathcal{L}_{global}$.
\begin{align}
    \begin{aligned}
        \mathcal{L}_{total}=\mathcal{L}_{mom}+\mathcal{L}_{high}+\lambda_{lg}(\mathcal{L}_{local}+\mathcal{L}_{global}),
    \end{aligned}
\end{align}
where $\lambda_{lg}$ is the coefficient of local-global regulators losses. $\mathcal{L}_{mom}$ and $\mathcal{L}_{high}$ are consistent with QD-DETR~\cite{moon2023query}.

\section{Experiment}
\begin{table*}[!ht]
    \centering
    \begin{tabular}{lcccccccc}
        \toprule  

    \multirow{4}{*}{\textbf{Method}} & \multirow{4}{*}{\textbf{Src}} & \multicolumn{5}{c}{\textbf{Moment Retrieval}}                           & \multicolumn{2}{c}{\textbf{HD}}                       \\ \cmidrule(r){3-7} \cmidrule(lr){8-9}
                            &                      & \multicolumn{2}{c}{R1} & \multicolumn{3}{c}{mAP} & \multicolumn{2}{c}{$\ge$Very Good} \\ \cmidrule(r){3-4} \cmidrule(lr){5-7} \cmidrule(lr){8-9}
                            &                      & @0.5       & @0.7      & @0.5   & @0.75  & Avg.  & mAP                   & HIT@1                \\
     \midrule
    BeautyThumb~\cite{DBLP:conf/cikm/SongRVJ16}      & V                    & -          & -         & -      & -      & -     & 14.36                 & 20.88                \\
    DVSE~\cite{DBLP:conf/cvpr/LiuMZCL15}            & V                    & -          & -         & -      & -      & -     & 18.75                 & 21.79                \\
    MCN~\cite{DBLP:conf/emnlp/HendricksWSSDR18}             & V                    & 11.41      & 2.72      & 24.94  & 8.22   & 10.67 & -                     & -                    \\
    CAL~\cite{DBLP:journals/corr/abs-1907-12763}             & V                    & 25.49      & 11.54     & 23.40   & 7.65   & 9.89  & -                     & -                    \\
    XML~\cite{DBLP:conf/eccv/LeiYBB20}            & V                    & 41.83      & 30.35     & 44.63  & 31.73  & 32.14 & 34.49                 & 55.25                \\
    XML+~\cite{lei2021detecting}            & V                    & 46.69      & 33.46     & 47.89  & 34.67  & 34.90  & 35.38                 & 55.06                \\
    MDETR~\cite{lei2021detecting}          & V                    & 52.89      & 33.02     & 54.82  & 29.40   & 30.73 & 35.69                 & 55.60                 \\
    QD-DETR~\cite{moon2023query}        & V                    & \underline{62.40}     & \underline{44.98}     & \underline{62.62}  & \underline{39.88}  & \underline{39.86} & \underline{38.64}                 & \underline{62.40}                 \\
    UniVTG~\cite{DBLP:journals/corr/abs-2307-16715}          & V                    & 58.86      & 40.86     & 57.60   & 35.59  & 35.47 & 38.20                  & 60.96                \\
    \rowcolor[rgb]{0.78,0.78,0.78}  TR-DETR                 & V                    & \textbf{64.66}     & \textbf{48.96}     & \textbf{63.98}  & \textbf{43.73}  & \textbf{42.62} & \textbf{39.91}                 & \textbf{63.42}                \\
    \midrule
    UMT~\cite{liu2022umt}            & V+A                  & 56.23      & 41.18     & 53.38  & 37.01  & 36.12 & 38.18                 & 59.99                \\
    QD-DETR~\cite{moon2023query}                & V+A                  & \underline{63.06}      & \underline{45.10}      & \underline{63.04}  & \underline{40.10}   & \underline{40.19} & \underline{39.04}                 & \underline{62.87}                \\
    \rowcolor[rgb]{0.78,0.78,0.78} TR-DETR                 & V+A                  & \textbf{65.05}      & \textbf{47.67}     & \textbf{64.87}  & \textbf{42.98}  & \textbf{43.10}  & \textbf{39.90}                  & \textbf{63.88}                \\
    \bottomrule
    \end{tabular}
    \caption{Experimental results on the QVHighlights \emph{test} set. HD represents the results of highlight detection. `V' and `A' represent using video and audio features, respectively. Bold letters indicate the best results, while underlined results are suboptimal.}
    \label{results_QV}
\end{table*}

\begin{table}[!ht]
    \centering
    \small
    \begin{tabular}{lccc}
        \toprule
        \textbf{Method} & \textbf{Feat} & R1@0.5 & R1@0.7 \\
        \midrule
        SAP~\cite{DBLP:conf/aaai/ChenJ19a} & VGG & 27.42 & 13.36 \\
        TripNet~\cite{DBLP:conf/bmvc/HahnKRG20} & VGG & 36.61 & 14.50 \\
        MAN~\cite{DBLP:conf/cvpr/ZhangDWWD19} & VGG & 41.24 & 20.54 \\
        2D-TAN~\cite{zhang2020learning} & VGG & 40.94 & 22.85 \\
        FVMR~\cite{li2021align} & VGG & 42.36 & 24.14 \\
        UMT†~\cite{liu2022umt} & VGG & 48.31 & 29.25 \\
        QD-DETR~\cite{moon2023query} & VGG & 52.77 & 31.13 \\
        QD-DETR†~\cite{moon2023query} & VGG & \textbf{55.51} & \textbf{34.17} \\
        \rowcolor[rgb]{0.78,0.78,0.78} TR-DETR & VGG & 53.47 & 30.81 \\
        \rowcolor[rgb]{0.78,0.78,0.78} TR-DETR† & VGG & \underline{54.49} & \underline{32.37} \\
        \midrule
        CTRL~\cite{gao2017tall} & C3D & 23.63 & 8.89 \\
        ACL~\cite{DBLP:conf/wacv/GeGCN19} & C3D & 30.48 & 12.20 \\
        MAN~\cite{DBLP:conf/cvpr/ZhangDWWD19} & C3D & 46.53 & 22.72 \\
        DEBUG~\cite{DBLP:conf/emnlp/LuCTLX19} & C3D & 37.39 & 17.69 \\
        VSLNet~\cite{zhang2021natural} & I3D & 47.31 & 30.19 \\
        QD-DETR~\cite{moon2023query} & I3D & \underline{50.67} & \underline{31.02} \\
        \rowcolor[rgb]{0.78,0.78,0.78} TR-DETR & I3D & \textbf{55.51} & \textbf{33.66} \\
        \midrule
        QD-DETR~\cite{moon2023query} & SF+C & 57.31 & 32.55 \\
        UniVTG~\cite{DBLP:journals/corr/abs-2307-16715} & SF+C & \textbf{58.01} & \textbf{35.65} \\
        \rowcolor[rgb]{0.78,0.78,0.78} TR-DETR & SF+C & \underline{57.61} & \underline{33.52} \\
        \bottomrule
    \end{tabular}
    \caption{Experimental results on the Charades-STA \emph{test} set. `†' represents using audio features.}
    \label{results_CHA}
\end{table}

\subsection{Datasets}

\begin{table*}[!ht]
    \centering
    \begin{tabular}{lcccccccccccc}
        \toprule
        \textbf{Method} & \textbf{Src} & VT & VU & GA & MS & PK & PR & FM & BK & BT & DS & Avg \\
        \midrule
        sLSTM~\cite{DBLP:conf/eccv/ZhangCSG16} & V & 41.1 & 46.2 & 46.3 & 47.7 & 44.8 & 46.1 & 45.2 & 40.6 & 47.1 & 45.5 & 45.1 \\
        SG~\cite{DBLP:journals/tmm/YuanTLF20} & V & 42.3 & 47.2 & 47.5 & 48.9 & 45.6 & 47.3 & 46.4 & 41.7 & 48.3 & 46.6 & 46.2 \\
        LIM-S~\cite{xiong2019less} & V & 55.9 & 42.9 & 61.2 & 54.0 & 60.3 & 47.5 & 43.2 & 66.3 & 69.1 & 62.6 & 56.3 \\
        Trailer~\cite{DBLP:conf/eccv/WangLPM20} & V & 61.3 & 54.6 & 65.7 & 60.8 & 59.1 & 70.1 & 58.2 & 64.7 & 65.6 & 68.1 & 62.8 \\
        SL-Module~\cite{DBLP:conf/iccv/XuWNZSW21} & V & 86.5 & 68.7 & 74.9 & \textbf{86.2} & 79.0 & 63.2 & 58.9 & 72.6 & 78.9 & 64.0 & 73.3 \\
        QD-DETR~\cite{moon2023query} & V & \underline{88.2} & \underline{87.4} & 85.6 & 85.0 & \underline{85.8} & 86.9 & \underline{76.4} & \underline{91.3} & \underline{89.2} & \underline{73.7} & \underline{85.0} \\
        UniVTG~\cite{DBLP:journals/corr/abs-2307-16715} & V & 83.9 & 85.1 & \underline{89.0} & 80.1 & 84.6 & \underline{87.0} & 70.9 & \textbf{91.7} & 73.5 & 69.3 & 81.0 \\
        \rowcolor[rgb]{0.78,0.78,0.78} TR-DETR & V & \textbf{89.3} & \textbf{93.0} & \textbf{94.3} & \underline{85.1} & \textbf{88.0} & \textbf{88.6} & \textbf{80.4} & \underline{91.3} & \textbf{89.5} & \textbf{81.6} & \textbf{88.1} \\
        \midrule
        MINI-Net~\cite{DBLP:conf/eccv/HongHLZ20} & V+A & 80.6 & 68.3 & 78.2 & 81.8 & 78.1 & 65.8 & 75.8 & 75.0 & 80.2 & 65.5 & 73.2 \\
        TCG~\cite{DBLP:conf/iccv/YeSGWBL021} & V+A & 85.0 & 71.4 & 81.9 & 78.6 & 80.2 & 75.5 & 71.6 & 77.3 & 78.6 & 68.1 & 76.8 \\
        Joint-VA~\cite{DBLP:conf/iccv/BadamdorjRWC21} & V+A & 83.7 & 57.3 & 78.5 & \underline{86.1} & 80.1 & 69.2 & 70.0 & 73.0 & \textbf{97.4} & 67.5 & 76.3 \\
        UMT~\cite{liu2022umt} & V+A & 87.5 & 81.5 & 88.2 & 78.8 & 81.4 & \underline{87.0} & 76.0 & 86.9 & 84.4 & \underline{79.6} & 83.1 \\
        QD-DETR~\cite{moon2023query} & V+A & \underline{87.6} & \underline{91.7} & \underline{90.2} & \textbf{88.3} & \underline{84.1} & \textbf{88.3} & \underline{78.7} & \underline{91.2} & 87.8 & 77.7 & \underline{86.6} \\
        \rowcolor[rgb]{0.78,0.78,0.78} TR-DETR & V+A & \textbf{90.6} & \textbf{92.4} & \textbf{91.7} & 81.3 & \textbf{86.9} & 85.5 & \textbf{79.8} & \textbf{93.4} & \underline{88.3} & \textbf{81.0} & \textbf{87.1} \\
        \bottomrule
    \end{tabular}
    \caption{Experimental results on the TVSum \emph{val} set. `V' and `A' represent using video and audio features, respectively.}
    \label{results_TV}
\end{table*}

\begin{figure*}[!ht]
\centering
\includegraphics[width=0.95\textwidth]{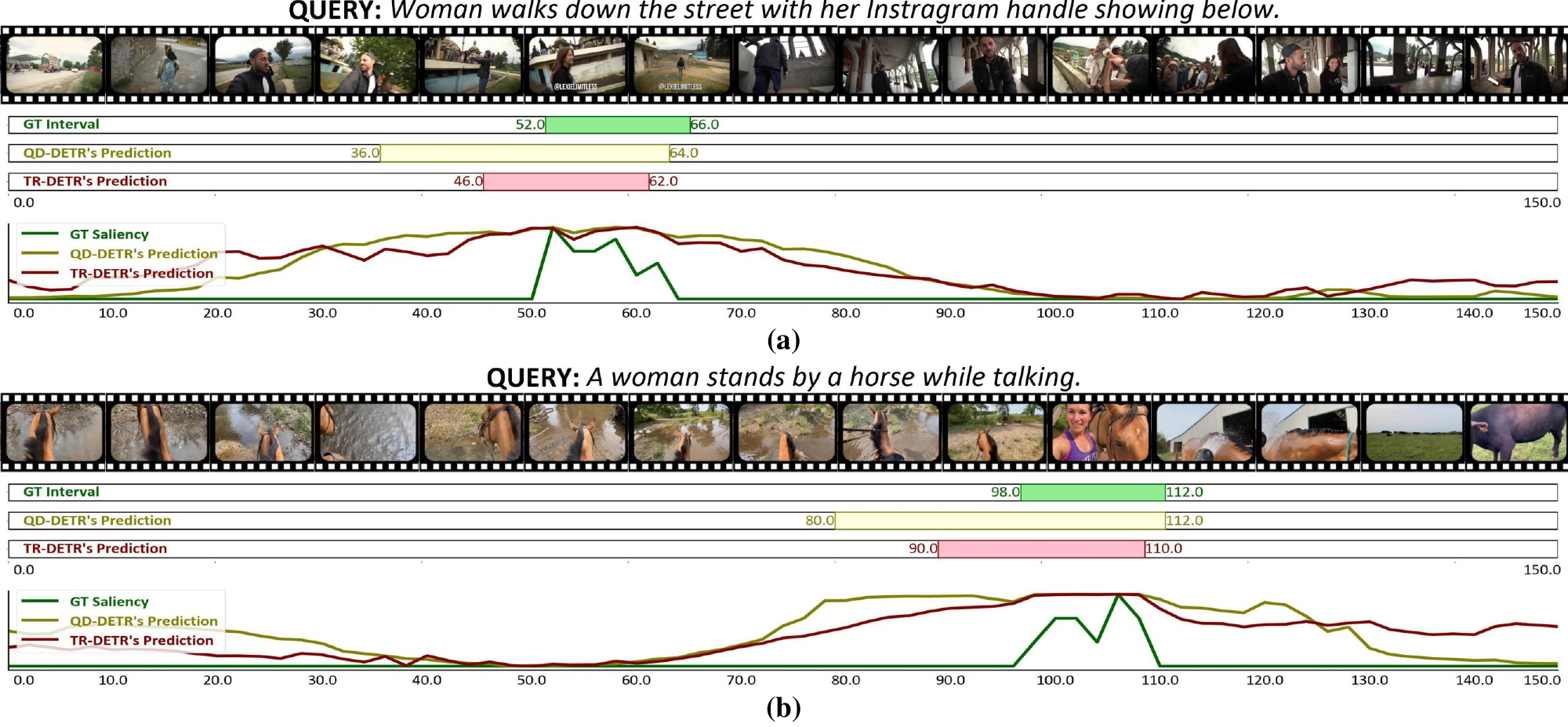} 
\caption{Qualitative results of TR-DETR on QVHighlights \emph{val} set.}
\label{qualative_QV}
\end{figure*}

\begin{figure*}[!ht]
\centering
\includegraphics[width=0.95\textwidth]{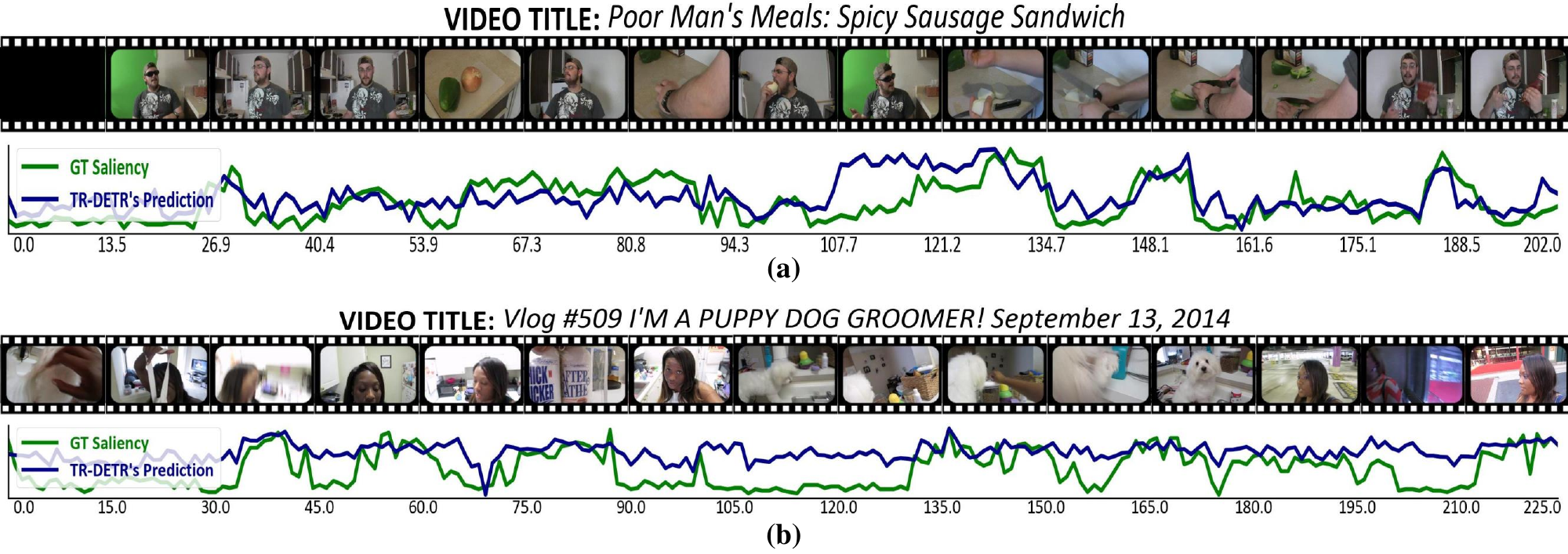} 
\caption{Qualitative results of TR-DETR on TVSum \emph{val} set.}
\label{qualative_TV}
\end{figure*}

\begin{figure}[!ht]
\centering
\includegraphics[width=0.45\textwidth]{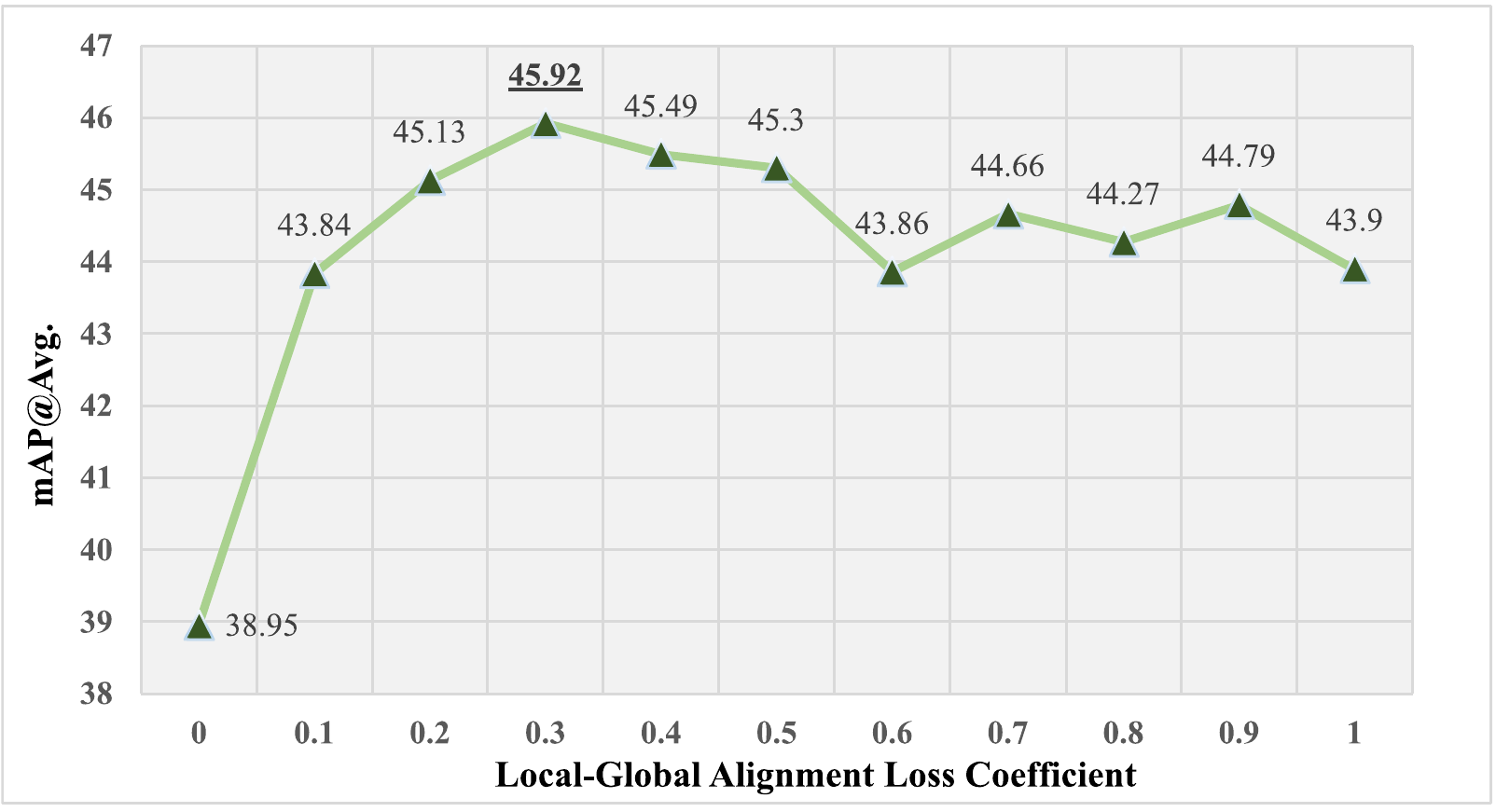} 
\caption{The impact of local-global alignment loss and $\lambda_{lg}$ based on QVHighlights \emph{val} set, introducing audio features. }
\label{cof_abi}
\end{figure}
QVHighlights dataset~\cite{lei2021detecting} comprises 10,148 content-rich videos from YouTube. Each video is accompanied by at least one manually annotated text query, where the highlight clips are located within the corresponding moment. The evaluation process of this dataset is particularly fair as the annotations of the test set are inaccessible. The prediction results of the model need to be uploaded to the QVHighlights server's CodaLab competition platform\footnote{https://codalab.lisn.upsaclay.fr/competitions/6937} for impartial performance assessment.

Charades-STA dataset~\cite{gao2017tall} contains 9,848 videos capturing daily indoor activities and 16,128 human-tagged query texts. Following QD-DETR\cite{moon2023query}, we allocate 12,408 samples for training while the remaining 3,720 samples are for testing.

TVSum dataset~\cite{song2015tvsum} is a benchmark dataset for HD. It contains 10 different categories of videos, and each category comprises 5 videos. To ensure consistency with QD-DETR~\cite{moon2023query},  80\% of the dataset is utilized for training and the remaining for testing.

\subsection{Metrics and Experimental Settings}
We use common metrics from recent studies like Moment-DETR, UMT, QD-DETR, and MH-DETR. For QVHighlights, we calculate Recall@1 with IoU $\in \{0.5, 0.7\}$ and mean average precision (mAP) with IoU $\in \{0.5, 0.75\}$. Following Lei \emph{et al.}~\cite{lei2021detecting}, we also uniformly sample 10 IoU thresholds from $\{0.5, 0.95\}$ to calculate mAP, and take the average as the average mAP metric. For highlight detection, we use mAP and HIT@1. Charades-STA involves Recall@1 with IoU $\in \{0.5, 0.7\}$, while for TVSum, top-5 mAP is the main metric.

In addition, we introduce implementation details and hyperparameters as follows. The hidden layer dimension $d$ is 256, and $\lambda_{lg}$ is set to 0.3. We use PANN~\cite{DBLP:journals/taslp/KongCIWWP20} trained on the AudioSet dataset~\cite{DBLP:conf/icassp/GemmekeEFJLMPR17} to extract audio features.
For QVHighlights, we use SlowFast~\cite{DBLP:journals/corr/abs-1812-03982} and CLIP to extract visual features and the text encoder in CLIP to extract textual features. The training phase involves 200 epochs, a batch size of 32, and a learning rate of 1e-4.
For TVSum, we use the I3D pre-trained on Kinetics-400 for visual features and CLIP for textual features. Training spans 2000 epochs with a batch size of 4 and a learning rate of 1e-3.
In Charades-STA, we extract visual features with VGG~\cite{DBLP:journals/corr/SimonyanZ14a}, I3D~\cite{DBLP:conf/cvpr/CarreiraZ17}, SlowFast, and CLIP, and use GLoVe~\cite{DBLP:conf/emnlp/PenningtonSM14}  for textual features. The training phase includes 100 epochs, a batch size of 8, and a learning rate of 1e-4. Moreover, all our experiments are conducted on Nvidia RTX 4090 and Gen Intel(R) Core(TM) i7-12700 CPU.

\subsection{Comparison with Other Methods}
Table~\ref{results_QV} reports the TR-DETR's performance on joint moment retrieval and highlight detection tasks. Meanwhile, Tables~\ref{results_CHA} and~\ref{results_TV} list the results of different methods on moment retrieval and highlight detection, respectively.

In Table~\ref{results_QV}, we evaluate the performance of moment retrieval and highlight detection simultaneously based on the QVHighlights dataset. For a fair comparison, we compare the performance with UniVTG~\cite{DBLP:journals/corr/abs-2307-16715} without pre-training. As shown in Table~\ref{results_QV}, our TR-DETR method outperforms the current best approach on all metrics.  Especially with visual features only, TR-DETR exhibits a significant increase in performance under more stringent metrics and high IOU thresholds.  Compared with previous methods, TR-DETR improves R1@0.7 and mAP@0.75 by 3.98\% and 3.75\%, respectively.
In addition, after introducing audio information, the performance of a few indicators decreases. This may be because the audio features are spliced directly behind the video features, causing misaligned multi-modal features to be combined and thus impairing modal interactions.

In Table~\ref{results_CHA}, we use VGG, C3D, and SF+C features to comprehensively evaluate the performance of TR-DETR on the Charades-STA dataset. For each feature of VGG, C3D and SF+C, we follow the data preparation settings of UMT~\cite{liu2022umt}, VSLNet~\cite{zhang2021natural}, and Moment-DETR~\cite{lei2021detecting}, respectively.  As shown in Table~\ref{results_CHA}, our TR-DETR shows comparable performance on VGG and SF+C features.  Also, performance on some metrics degrades with the introduction of audio, possibly due to insufficient modal interaction. Compared with using only VGG features, the performance of the proposed method is slightly different from UniVTG when using SF+C features. We believe the reasons are as follows: semantic information of features extracted by different-scale feature extractors (\emph{e.g.} VGG and PANN) varies greatly. In our method, the local-global multi-modal alignment module is used to force the alignment of the visual features of VGG, the audio features of PANN, and the text features of GLoVe, which is challenging and results in relatively weak performance. However, when text, visual and audio features are all derived from large models, such as CLIP, our method shows excellent performance on the  QVHighlights dataset.

Consistent with previous work on highlight detection, we evaluate the performance of the proposed TR-DETR on each video category and calculate the top-5 mAP scores. The results are shown in Table~\ref{results_TV}. In addition, to comprehensively evaluate the overall performance of TR-DETR, we calculate the average value of top-5 mAP on 10 categories. The proposed TR-DETR exceeds the previous method by approximately 3.1\% when using only video features, which demonstrates the powerful performance of TR-DETR in solving HD alone.

\begin{table*}[t]
    \centering
    \small
    \begin{tabular}{ccccclllllll}
        \toprule
        \multirow{4}{*}{\textbf{Setting}} & \multirow{4}{*}{LGAM} & \multirow{4}{*}{VFR} & \multirow{4}{*}{MR2HD} & \multirow{4}{*}{HD2MR} & \multicolumn{5}{c}{\textbf{Moment Retrieval}}                           & \multicolumn{2}{c}{\textbf{HD}}                        \\ \cmidrule(r){6-10} \cmidrule(lr){11-12}
                             &                     &                              &                        &                        & \multicolumn{2}{c}{R1} & \multicolumn{3}{c}{mAP} & \multicolumn{2}{c}{$\ge$Very Good} \\ \cmidrule(r){6-7} \cmidrule(r){8-10} \cmidrule(lr){11-12}
                             &                     &                              &                        &                        & @0.5       & @0.7      & @0.5   & @0.75  & Avg.  & mAP                    & HIT@1                \\
        \midrule
        (a)                      &                     &                              &                        &                        & 57.72      & 42.35     & 59.10  & 38.16  & 38.03 & 36.76                  & 57.44                \\
        \midrule
        (b)                      & \checkmark                 &                              &                        &                        & 63.10       & 44.97     & 63.13  & 40.22  & 40.47 & 39.92                  & 63.87                \\
        (c)                      &                     & \checkmark                             &                        &                        & 64.19      & 47.61     & 63.50 & 42.90  & 41.74 & 39.71                  & 64.13                \\
        (d)                      &                     &                              & \checkmark                       &                        & 58.39      & 42.71     & 59.28  & 39.19  & 38.76 & 37.80                   & 58.8                 \\
        (e)                      &                     &                              &                        & \checkmark                       & 59.61      & 42.26     & 60.91  & 39.28  & 39.26 & 37.67                  & 58.45                \\
        \midrule
        (f)                      &                     &                              & \checkmark                       & \checkmark                       & 59.81      & 44.71     & 60.25  & 39.33  & 39.80  & 37.86                  & 57.94                \\
        (g)                      & \checkmark                    &                              & \checkmark                       & \checkmark                       & 62.13      & 47.16     & 62.00     & 42.79  & 41.21 & 39.76                  & 62.65                \\
        (h)                      &                     & \checkmark                             & \checkmark                       & \checkmark                       & 63.23      & 46.90      & 63.30   & 42.47  & 41.64 & 38.12                  & 59.55                \\
        (i)                      & \checkmark                    & \checkmark                             &                        &                        & 66.32      & 50.71     & 65.71  & 44.82  & 43.95 & 40.35                  & 64.90                 \\
        \midrule
        (j)                      & \checkmark                    & \checkmark                             & \checkmark                       & \checkmark                       & 67.10       & 51.48     & 66.27  & 46.42  & 45.09 & 40.55                  & 64.77               \\
        \bottomrule
    \end{tabular}
     \caption{Comparison with the baseline (Moment-DETR with cross-attention module and DAB-DETR's decoder~\cite{DBLP:conf/iclr/LiuLZYQSZZ22}) with different module combinations on QVHighlights \emph{val} set.  LGAM represents the local-global alignment module, and VFR is the visual feature refinement module.}
     \label{results_abi}
\end{table*}

\subsection{Visualization}

In Figures~\ref{qualative_QV} and~\ref{qualative_TV}, we visualize the qualitative analysis results of TR-DETR on the QVHighlights and TVSum datasets, respectively. In Figure~\ref{qualative_QV}, compared with QD-DETR, TR-DETR shows more reasonable and accurate results in terms of retrieved accuracy and highlight score distribution. In Figure~\ref{qualative_TV} a), the proposed TR-DETR can accurately fit the highlight score distribution. We believe that these performance improvements are due to the combination of the proposed modules. In addition, in Figure~\ref{qualative_TV} b), it may be that the model only noticed the concept of ``puppy dog", resulting in unreasonable high highlight scores in the middle of the result.

\subsection{Ablation}

To verify the effect of each module in the proposed TR-DETR, we conduct a comprehensive ablation experiment, and the results are listed in Table~\ref{results_abi}. Settings (b) to (e) show the performance of each component on the baseline model compared to setting (a). Setting (f) demonstrates the existence of task reciprocity. Compared with setting (c), the reason for the performance degradation in setting (h) may be the semantic mismatch between modalities, resulting in mutual degradation of tasks. Setting (i) shows the huge performance improvement of the proposed local-global alignment loss combined with visual feature refinement.

To further verify the effect of the proposed local-global alignment loss, we also conduct ablation experiments on its coefficients. As shown in Figure~\ref{cof_abi}, after adding the local global regularization term, the model's performance has been significantly improved by about 5\%. In addition, as the value of the hyperparameter $\lambda_{lg}$ gradually increases, the performance improvement becomes more significant. When $\lambda_{lg}$ is set to 0.3, the model performance reaches its peak and then begins to decline slowly. Comparing the hyperparameter values of 0 and 0.3, the model performance has been improved by about 7\% in total, confirming the significant role of the local-global alignment regulators.

\section{Conclusion}

This paper proposes a TR-DETR to explore the reciprocity between HD and MR tasks.
First, local-global alignment regulators are designed to align visual and textual features. Then, a visual feature refinement module is constructed to obtain discriminative joint features. Finally, a task-reciprocal module is proposed to inject highlight score information into the moment retrieval pipeline and optimize highlight score prediction by utilizing retrieved moments.  Extensive experiments on several datasets demonstrate the effectiveness of TR-DETR.
However, TR-DETR cannot efficiently utilize data from the audio modality. In the future, we will study novel multi-modal feature interaction networks to coordinate information from multiple modalities.


\section{Acknowledgments}
This work was supported in part by National Natural Science Foundation of China under Grant 62201222 and 62377026, in part by Hubei Provincial Natural Science Foundation of China under Grant 2022CFB954, in part by Knowledge Innovation Program of Wuhan-Shuguang Project under Grant 2023010201020377 and 2023010201020382, in part by self-determined research funds of CCNU from the colleges' basic research and operation of MOE under Grant CCNU22QN014, CCNU22JC007 and CCNU22XJ034, and in part by Hubei Provincial Key Laboratory of Artificial Intelligence and Smart Learning NO. 2023AISL003 and 2023AISL010.

\bibliography{aaai24}

\begin{thebibliography}{60}
\providecommand{\natexlab}[1]{#1}

\bibitem[{Badamdorj et~al.(2021)Badamdorj, Rochan, Wang, and Cheng}]{DBLP:conf/iccv/BadamdorjRWC21}
Badamdorj, T.; Rochan, M.; Wang, Y.; and Cheng, L. 2021.
\newblock Joint Visual and Audio Learning for Video Highlight Detection.
\newblock In \emph{IEEE ICCV}, 8107--8117.

\bibitem[{Carion et~al.(2020)Carion, Massa, Synnaeve, Usunier, Kirillov, and Zagoruyko}]{DBLP:conf/eccv/CarionMSUKZ20}
Carion, N.; Massa, F.; Synnaeve, G.; Usunier, N.; Kirillov, A.; and Zagoruyko, S. 2020.
\newblock End-to-End Object Detection with Transformers.
\newblock In \emph{ECCV}, 213--229. Springer.

\bibitem[{Carreira and Zisserman(2017)}]{DBLP:conf/cvpr/CarreiraZ17}
Carreira, J.; and Zisserman, A. 2017.
\newblock Quo Vadis, Action Recognition? {A} New Model and the Kinetics Dataset.
\newblock In \emph{IEEE CVPR}, 4724--4733.

\bibitem[{Chen and Jiang(2019)}]{DBLP:conf/aaai/ChenJ19a}
Chen, S.; and Jiang, Y. 2019.
\newblock Semantic Proposal for Activity Localization in Videos via Sentence Query.
\newblock In \emph{AAAI}, 8199--8206.

\bibitem[{Chung et~al.(2014)Chung, Gulcehre, Cho, and Bengio}]{chung2014empirical}
Chung, J.; Gulcehre, C.; Cho, K.; and Bengio, Y. 2014.
\newblock Empirical evaluation of gated recurrent neural networks on sequence modeling.
\newblock In \emph{NeurIPS 2014 Workshop on Deep Learning}.

\bibitem[{Escorcia et~al.(2019)Escorcia, Soldan, Sivic, Ghanem, and Russell}]{DBLP:journals/corr/abs-1907-12763}
Escorcia, V.; Soldan, M.; Sivic, J.; Ghanem, B.; and Russell, B.~C. 2019.
\newblock Temporal Localization of Moments in Video Collections with Natural Language.
\newblock \emph{CoRR}, abs/1907.12763.

\bibitem[{Feichtenhofer et~al.(2019)Feichtenhofer, Fan, Malik, and He}]{DBLP:journals/corr/abs-1812-03982}
Feichtenhofer, C.; Fan, H.; Malik, J.; and He, K. 2019.
\newblock SlowFast Networks for Video Recognition.
\newblock In \emph{IEEE ICCV}, 6201--6210.

\bibitem[{Foo et~al.(2023)Foo, Gong, Fan, and Liu}]{Foo_2023_CVPR}
Foo, L.~G.; Gong, J.; Fan, Z.; and Liu, J. 2023.
\newblock System-Status-Aware Adaptive Network for Online Streaming Video Understanding.
\newblock In \emph{IEEE CVPR}, 10514--10523.

\bibitem[{Gao et~al.(2017)Gao, Sun, Yang, and Nevatia}]{gao2017tall}
Gao, J.; Sun, C.; Yang, Z.; and Nevatia, R. 2017.
\newblock {TALL:} Temporal Activity Localization via Language Query.
\newblock In \emph{IEEE ICCV}, 5277--5285.

\bibitem[{Ge et~al.(2019)Ge, Gao, Chen, and Nevatia}]{DBLP:conf/wacv/GeGCN19}
Ge, R.; Gao, J.; Chen, K.; and Nevatia, R. 2019.
\newblock {MAC:} Mining Activity Concepts for Language-Based Temporal Localization.
\newblock In \emph{IEEE WACV}, 245--253.

\bibitem[{Gemmeke et~al.(2017)Gemmeke, Ellis, Freedman, Jansen, Lawrence, Moore, Plakal, and Ritter}]{DBLP:conf/icassp/GemmekeEFJLMPR17}
Gemmeke, J.~F.; Ellis, D. P.~W.; Freedman, D.; Jansen, A.; Lawrence, W.; Moore, R.~C.; Plakal, M.; and Ritter, M. 2017.
\newblock Audio Set: An ontology and human-labeled dataset for audio events.
\newblock In \emph{IEEE ICASSP}, 776--780.

\bibitem[{Ghosh et~al.(2019)Ghosh, Agarwal, Parekh, and Hauptmann}]{ghosh2019excl}
Ghosh, S.; Agarwal, A.; Parekh, Z.; and Hauptmann, A. 2019.
\newblock {E}x{CL}: {E}xtractive {C}lip {L}ocalization {U}sing {N}atural {L}anguage {D}escriptions.
\newblock In \emph{NAACL}, 1984--1990. Minneapolis, Minnesota: ACL.

\bibitem[{Ging et~al.(2020)Ging, Zolfaghari, Pirsiavash, and Brox}]{ging2020coot}
Ging, S.; Zolfaghari, M.; Pirsiavash, H.; and Brox, T. 2020.
\newblock {COOT:} Cooperative Hierarchical Transformer for Video-Text Representation Learning.
\newblock In \emph{NeurIPS}.

\bibitem[{Guo et~al.(2022)Guo, Zhao, Jin, Wang, Liu, and Yu}]{DBLP:journals/tmm/GuoZJWLY22}
Guo, Z.; Zhao, Z.; Jin, W.; Wang, D.; Liu, R.; and Yu, J. 2022.
\newblock TaoHighlight: Commodity-Aware Multi-Modal Video Highlight Detection in E-Commerce.
\newblock \emph{IEEE TMM}, 24: 2606--2616.

\bibitem[{Hahn et~al.(2020)Hahn, Kadav, Rehg, and Graf}]{DBLP:conf/bmvc/HahnKRG20}
Hahn, M.; Kadav, A.; Rehg, J.~M.; and Graf, H.~P. 2020.
\newblock Tripping through time: Efficient Localization of Activities in Videos.
\newblock In \emph{BMVC}.

\bibitem[{Hendricks et~al.(2018)Hendricks, Wang, Shechtman, Sivic, Darrell, and Russell}]{DBLP:conf/emnlp/HendricksWSSDR18}
Hendricks, L.~A.; Wang, O.; Shechtman, E.; Sivic, J.; Darrell, T.; and Russell, B.~C. 2018.
\newblock Localizing Moments in Video with Temporal Language.
\newblock In \emph{EMNLP}, 1380--1390. ACL.

\bibitem[{Hong et~al.(2020)Hong, Huang, Li, and Zheng}]{DBLP:conf/eccv/HongHLZ20}
Hong, F.; Huang, X.; Li, W.; and Zheng, W. 2020.
\newblock MINI-Net: Multiple Instance Ranking Network for Video Highlight Detection.
\newblock In \emph{ECCV}, 345--360. Springer.

\bibitem[{Kong et~al.(2020)Kong, Cao, Iqbal, Wang, Wang, and Plumbley}]{DBLP:journals/taslp/KongCIWWP20}
Kong, Q.; Cao, Y.; Iqbal, T.; Wang, Y.; Wang, W.; and Plumbley, M.~D. 2020.
\newblock PANNs: Large-Scale Pretrained Audio Neural Networks for Audio Pattern Recognition.
\newblock \emph{IEEE TASLP}, 28: 2880--2894.

\bibitem[{Kudi and Namboodiri(2017)}]{kudi2017words}
Kudi, S.; and Namboodiri, A.~M. 2017.
\newblock Words speak for actions: Using text to find video highlights.
\newblock In \emph{ACPR}, 322--327. IEEE.

\bibitem[{Lei, Berg, and Bansal(2021)}]{lei2021detecting}
Lei, J.; Berg, T.~L.; and Bansal, M. 2021.
\newblock Detecting Moments and Highlights in Videos via Natural Language Queries.
\newblock In \emph{NeurIPS}, 11846--11858.

\bibitem[{Lei et~al.(2020)Lei, Yu, Berg, and Bansal}]{DBLP:conf/eccv/LeiYBB20}
Lei, J.; Yu, L.; Berg, T.~L.; and Bansal, M. 2020.
\newblock {TVR:} {A} Large-Scale Dataset for Video-Subtitle Moment Retrieval.
\newblock In \emph{ECCV}, 447--463. Springer.

\bibitem[{Li et~al.(2021)Li, Selvaraju, Gotmare, Joty, Xiong, and Hoi}]{li2021align}
Li, J.; Selvaraju, R.~R.; Gotmare, A.; Joty, S.~R.; Xiong, C.; and Hoi, S.~C. 2021.
\newblock Align before Fuse: Vision and Language Representation Learning with Momentum Distillation.
\newblock In \emph{NeurIPS}, 9694--9705.

\bibitem[{Li et~al.(2022)Li, Xie, Qian, Zhu, Tang, Wu, Yang, Zhuang, and Wang}]{DBLP:conf/cvpr/0006XQZT0YZW22}
Li, J.; Xie, J.; Qian, L.; Zhu, L.; Tang, S.; Wu, F.; Yang, Y.; Zhuang, Y.; and Wang, X.~E. 2022.
\newblock Compositional Temporal Grounding with Structured Variational Cross-Graph Correspondence Learning.
\newblock In \emph{IEEE CVPR}, 3022--3031.

\bibitem[{Lin et~al.(2023)Lin, Zhang, Chen, Pramanick, Gao, Wang, Yan, and Shou}]{DBLP:journals/corr/abs-2307-16715}
Lin, K.~Q.; Zhang, P.; Chen, J.; Pramanick, S.; Gao, D.; Wang, A.~J.; Yan, R.; and Shou, M.~Z. 2023.
\newblock UniVTG: Towards Unified Video-Language Temporal Grounding.
\newblock \emph{CoRR}, abs/2307.16715.

\bibitem[{Liu et~al.(2022{\natexlab{a}})Liu, Li, Zhang, Yang, Qi, Su, Zhu, and Zhang}]{DBLP:conf/iclr/LiuLZYQSZZ22}
Liu, S.; Li, F.; Zhang, H.; Yang, X.; Qi, X.; Su, H.; Zhu, J.; and Zhang, L. 2022{\natexlab{a}}.
\newblock {DAB-DETR:} Dynamic Anchor Boxes are Better Queries for {DETR}.
\newblock In \emph{ICLR}.

\bibitem[{Liu et~al.(2015)Liu, Mei, Zhang, Che, and Luo}]{DBLP:conf/cvpr/LiuMZCL15}
Liu, W.; Mei, T.; Zhang, Y.; Che, C.; and Luo, J. 2015.
\newblock Multi-task deep visual-semantic embedding for video thumbnail selection.
\newblock In \emph{IEEE CVPR}, 3707--3715.

\bibitem[{Liu et~al.(2022{\natexlab{b}})Liu, Li, Wu, Chen, Shan, and Qie}]{liu2022umt}
Liu, Y.; Li, S.; Wu, Y.; Chen, C.-W.; Shan, Y.; and Qie, X. 2022{\natexlab{b}}.
\newblock Umt: Unified multi-modal transformers for joint video moment retrieval and highlight detection.
\newblock In \emph{IEEE CVPR}, 3042--3051.

\bibitem[{Lu et~al.(2019)Lu, Chen, Tan, Li, and Xiao}]{DBLP:conf/emnlp/LuCTLX19}
Lu, C.; Chen, L.; Tan, C.; Li, X.; and Xiao, J. 2019.
\newblock {DEBUG:} {A} Dense Bottom-Up Grounding Approach for Natural Language Video Localization.
\newblock In \emph{EMNLP-IJCNLP}, 5143--5152. ACL.

\bibitem[{Luo et~al.(2020)Luo, Ji, Shi, Huang, Duan, Li, Li, Bharti, and Zhou}]{luo2020univl}
Luo, H.; Ji, L.; Shi, B.; Huang, H.; Duan, N.; Li, T.; Li, J.; Bharti, T.; and Zhou, M. 2020.
\newblock Univl: A unified video and language pre-training model for multimodal understanding and generation.
\newblock \emph{ArXiv preprint ArXiv:2002.06353}.

\bibitem[{Miech et~al.(2020)Miech, Alayrac, Smaira, Laptev, Sivic, and Zisserman}]{miech2020end}
Miech, A.; Alayrac, J.; Smaira, L.; Laptev, I.; Sivic, J.; and Zisserman, A. 2020.
\newblock End-to-End Learning of Visual Representations From Uncurated Instructional Videos.
\newblock In \emph{IEEE CVPR}, 9876--9886.

\bibitem[{Molino and Gygli(2018)}]{DBLP:conf/mm/MolinoG18}
Molino, A. G.~D.; and Gygli, M. 2018.
\newblock PHD-GIFs: Personalized Highlight Detection for Automatic {GIF} Creation.
\newblock In \emph{MM}, 600--608. ACM.

\bibitem[{Moon et~al.(2023)Moon, Hyun, Park, Park, and Heo}]{moon2023query}
Moon, W.; Hyun, S.; Park, S.; Park, D.; and Heo, J.-P. 2023.
\newblock Query-dependent video representation for moment retrieval and highlight detection.
\newblock In \emph{IEEE CVPR}, 23023--23033.

\bibitem[{Mun, Cho, and Han(2020)}]{mun2020local}
Mun, J.; Cho, M.; and Han, B. 2020.
\newblock Local-Global Video-Text Interactions for Temporal Grounding.
\newblock In \emph{IEEE CVPR}, 10807--10816. {IEEE}.

\bibitem[{Pennington, Socher, and Manning(2014)}]{DBLP:conf/emnlp/PenningtonSM14}
Pennington, J.; Socher, R.; and Manning, C.~D. 2014.
\newblock Glove: Global Vectors for Word Representation.
\newblock In \emph{EMNLP}, 1532--1543. ACL.

\bibitem[{Radford et~al.(2021)Radford, Kim, Hallacy, Ramesh, Goh, Agarwal, Sastry, Askell, Mishkin, Clark, Krueger, and Sutskever}]{DBLP:conf/icml/RadfordKHRGASAM21}
Radford, A.; Kim, J.~W.; Hallacy, C.; Ramesh, A.; Goh, G.; Agarwal, S.; Sastry, G.; Askell, A.; Mishkin, P.; Clark, J.; Krueger, G.; and Sutskever, I. 2021.
\newblock Learning Transferable Visual Models From Natural Language Supervision.
\newblock In \emph{ICML}, volume 139, 8748--8763.

\bibitem[{Scarselli et~al.(2008)Scarselli, Gori, Tsoi, Hagenbuchner, and Monfardini}]{scarselli2008graph}
Scarselli, F.; Gori, M.; Tsoi, A.~C.; Hagenbuchner, M.; and Monfardini, G. 2008.
\newblock The graph neural network model.
\newblock \emph{TNNLS}, 20(1): 61--80.

\bibitem[{Simonyan and Zisserman(2015)}]{DBLP:journals/corr/SimonyanZ14a}
Simonyan, K.; and Zisserman, A. 2015.
\newblock Very Deep Convolutional Networks for Large-Scale Image Recognition.
\newblock In \emph{ICLR}.

\bibitem[{Song et~al.(2016)Song, Redi, Vallmitjana, and Jaimes}]{DBLP:conf/cikm/SongRVJ16}
Song, Y.; Redi, M.; Vallmitjana, J.; and Jaimes, A. 2016.
\newblock To Click or Not To Click: Automatic Selection of Beautiful Thumbnails from Videos.
\newblock In \emph{ACM CIKM}, 659--668.

\bibitem[{Song et~al.(2015)Song, Vallmitjana, Stent, and Jaimes}]{song2015tvsum}
Song, Y.; Vallmitjana, J.; Stent, A.; and Jaimes, A. 2015.
\newblock TVSum: Summarizing web videos using titles.
\newblock In \emph{IEEE CVPR}, 5179--5187.

\bibitem[{Sun et~al.(2020)Sun, Baradel, Murphy, and Schmid}]{sun2019learning}
Sun, C.; Baradel, F.; Murphy, K.; and Schmid, C. 2020.
\newblock Learning Video Representations using Contrastive Bidirectional Transformer.

\bibitem[{Vaswani et~al.(2017)Vaswani, Shazeer, Parmar, Uszkoreit, Jones, Gomez, Kaiser, and Polosukhin}]{vaswani2017attention}
Vaswani, A.; Shazeer, N.; Parmar, N.; Uszkoreit, J.; Jones, L.; Gomez, A.~N.; Kaiser, {\L}.; and Polosukhin, I. 2017.
\newblock Attention is all you need.
\newblock In \emph{NeurIPS}, 5998--6008.

\bibitem[{Wang et~al.(2020)Wang, Liu, Puri, and Metaxas}]{DBLP:conf/eccv/WangLPM20}
Wang, L.; Liu, D.; Puri, R.; and Metaxas, D.~N. 2020.
\newblock Learning Trailer Moments in Full-Length Movies with Co-Contrastive Attention.
\newblock In \emph{ECCV}, 300--316. Springer.

\bibitem[{Wang et~al.(2022)Wang, Wang, Wu, Li, and Wu}]{DBLP:conf/aaai/00010WLW22}
Wang, Z.; Wang, L.; Wu, T.; Li, T.; and Wu, G. 2022.
\newblock Negative Sample Matters: {A} Renaissance of Metric Learning for Temporal Grounding.
\newblock In \emph{AAAI}, 2613--2623. {AAAI} Press.

\bibitem[{Xiong et~al.(2019)Xiong, Kalantidis, Ghadiyaram, and Grauman}]{xiong2019less}
Xiong, B.; Kalantidis, Y.; Ghadiyaram, D.; and Grauman, K. 2019.
\newblock Less Is More: Learning Highlight Detection From Video Duration.
\newblock In \emph{IEEE CVPR}, 1258--1267.

\bibitem[{Xiong, Zhong, and Socher(2017)}]{xiong2016dynamic}
Xiong, C.; Zhong, V.; and Socher, R. 2017.
\newblock Dynamic Coattention Networks For Question Answering.
\newblock In \emph{ICLR}.

\bibitem[{Xiong and Wang(2023)}]{xiong2023dual}
Xiong, Z.; and Wang, H. 2023.
\newblock Dual-Stream Multimodal Learning for Topic-Adaptive Video Highlight Detection.
\newblock In \emph{ICMR}, 272--279. ACM.

\bibitem[{Xu et~al.(2019)Xu, He, Plummer, Sigal, Sclaroff, and Saenko}]{xu2019multilevel}
Xu, H.; He, K.; Plummer, B.~A.; Sigal, L.; Sclaroff, S.; and Saenko, K. 2019.
\newblock Multilevel Language and Vision Integration for Text-to-Clip Retrieval.
\newblock In \emph{AAAI}, 9062--9069.

\bibitem[{Xu et~al.(2021)Xu, Wang, Ni, Zhu, Sun, and Wang}]{DBLP:conf/iccv/XuWNZSW21}
Xu, M.; Wang, H.; Ni, B.; Zhu, R.; Sun, Z.; and Wang, C. 2021.
\newblock Cross-category Video Highlight Detection via Set-based Learning.
\newblock In \emph{IEEE ICCV}, 7950--7959.

\bibitem[{Xu, Zhu, and Clifton(2022)}]{10123038}
Xu, P.; Zhu, X.; and Clifton, D.~A. 2022.
\newblock Multimodal Learning With Transformers: A Survey.
\newblock \emph{IEEE TPAMI}, 45: 12113--12132.

\bibitem[{Xu et~al.(2023)Xu, Sun, Li, Shi, Zhu, and Du}]{xu2023mh}
Xu, Y.; Sun, Y.; Li, Y.; Shi, Y.; Zhu, X.; and Du, S. 2023.
\newblock MH-DETR: Video Moment and Highlight Detection with Cross-modal Transformer.
\newblock \emph{ArXiv preprint ArXiv:2305.00355}.

\bibitem[{Yan et~al.(2023)Yan, Chen, Song, and Zhu}]{yan2023CAAI}
Yan, Z.; Chen, Y.; Song, J.; and Zhu, J. 2023.
\newblock Multimodal feature fusion based on object relation for video captioning.
\newblock \emph{CAAI TRIT}, 8(1): 247--259.

\bibitem[{Ye et~al.(2021)Ye, Shen, Gao, Wang, Bi, Li, and Yang}]{DBLP:conf/iccv/YeSGWBL021}
Ye, Q.; Shen, X.; Gao, Y.; Wang, Z.; Bi, Q.; Li, P.; and Yang, G. 2021.
\newblock Temporal Cue Guided Video Highlight Detection with Low-Rank Audio-Visual Fusion.
\newblock In \emph{IEEE ICCV}, 7930--7939.

\bibitem[{Yuan et~al.(2020)Yuan, Tay, Li, and Feng}]{DBLP:journals/tmm/YuanTLF20}
Yuan, L.; Tay, F. E.~H.; Li, P.; and Feng, J. 2020.
\newblock Unsupervised Video Summarization With Cycle-Consistent Adversarial {LSTM} Networks.
\newblock \emph{IEEE TMM}, 22(10): 2711--2722.

\bibitem[{Yuan, Ma, and Zhu(2019)}]{DBLP:conf/mm/YuanM019}
Yuan, Y.; Ma, L.; and Zhu, W. 2019.
\newblock Sentence Specified Dynamic Video Thumbnail Generation.
\newblock In \emph{MM}, 2332--2340. ACM.

\bibitem[{Zhang et~al.(2022)Zhang, Yang, Jiang, and Zhou}]{zhang2022video}
Zhang, B.; Yang, C.; Jiang, B.; and Zhou, X. 2022.
\newblock Video Moment Retrieval with Hierarchical Contrastive Learning.
\newblock In \emph{MM}, 346--355. ACM.

\bibitem[{Zhang et~al.(2019)Zhang, Dai, Wang, Wang, and Davis}]{DBLP:conf/cvpr/ZhangDWWD19}
Zhang, D.; Dai, X.; Wang, X.; Wang, Y.; and Davis, L.~S. 2019.
\newblock {MAN:} Moment Alignment Network for Natural Language Moment Retrieval via Iterative Graph Adjustment.
\newblock In \emph{IEEE CVPR}, 1247--1257.

\bibitem[{Zhang et~al.(2021)Zhang, Sun, Jing, Zhen, Zhou, and Goh}]{zhang2021natural}
Zhang, H.; Sun, A.; Jing, W.; Zhen, L.; Zhou, J.~T.; and Goh, R. S.~M. 2021.
\newblock Natural language video localization: A revisit in span-based question answering framework.
\newblock \emph{IEEE transactions on pattern analysis and machine intelligence}, 44(8): 4252--4266.

\bibitem[{Zhang et~al.(2023)Zhang, Sun, Jing, and Zhou}]{DBLP:journals/pami/ZhangSJZ23}
Zhang, H.; Sun, A.; Jing, W.; and Zhou, J.~T. 2023.
\newblock Temporal Sentence Grounding in Videos: {A} Survey and Future Directions.
\newblock \emph{IEEE TPAMI}, 45(8): 10443--10465.

\bibitem[{Zhang et~al.(2016)Zhang, Chao, Sha, and Grauman}]{DBLP:conf/eccv/ZhangCSG16}
Zhang, K.; Chao, W.; Sha, F.; and Grauman, K. 2016.
\newblock Video Summarization with Long Short-Term Memory.
\newblock In \emph{ECCV}, 766--782. Springer.

\bibitem[{Zhang et~al.(2020)Zhang, Peng, Fu, and Luo}]{zhang2020learning}
Zhang, S.; Peng, H.; Fu, J.; and Luo, J. 2020.
\newblock Learning 2D Temporal Adjacent Networks for Moment Localization with Natural Language.
\newblock In \emph{AAAI}, 12870--12877.

\end{thebibliography}

\end{document}